\documentclass[10pt, conference, compsocconf]{IEEEtran}
\ifCLASSINFOpdf
\else
\fi
\usepackage{amsmath}
\usepackage{gensymb}
\usepackage{graphicx}
\usepackage{subcaption}
\usepackage{nameref}
\usepackage{varioref}
\usepackage{hyperref}
\usepackage{cleveref}
\begin{document}
%
\title{Sleep-deprived Fatigue Pattern Analysis using Large-Scale Selfies from Social Media}


\author{\IEEEauthorblockN{Xuefeng Peng}
\IEEEauthorblockA{Computer Science Department\\
University of Rochester\\
Rochester, USA\\
xpeng4@u.rochester.edu}
\and
\IEEEauthorblockN{Jiebo Luo}
\IEEEauthorblockA{Computer Science Department\\
University of Rochester\\
Rochester, USA\\
jiebo.luo@rochester.edu}
\and
\IEEEauthorblockN{Catherine Glenn}
\IEEEauthorblockA{Department of Psychiatry\\
University of Rochester\\
Rochester, USA\\
catherine.glenn@rochester.edu}
\and
\IEEEauthorblockN{Li-Kai Chi, Jingyao Zhan}
\IEEEauthorblockA{Computer Science Department\\
University of Rochester\\
Rochester, USA\\
\{lchi3, jzhan\}@u.rochester.edu}
}


%


\maketitle

\begin{abstract}
The complexities of fatigue have drawn much attention from researchers across various disciplines. Short-term fatigue may cause safety issue while driving; thus, dynamic systems were designed to track driver fatigue. Long-term fatigue could lead to chronic syndromes, and eventually affect individual’s physical and psychological health. Traditional methodologies of evaluating fatigue not only require sophisticated equipment but also consume enormous time. In this paper, we attempt to develop a novel and efficient method to predict individual's fatigue rate by scrutinizing human facial cues. Our goal is to predict fatigue rate based on a selfie. To associate the fatigue rate with user behaviors, we have collected nearly 1-million timeline posts from 10,480 users on Instagram. We first detect all the faces and identify their demographics using automatic algorithms. Next, we investigate the fatigue distribution by weekday over different age, gender, and ethnic groups. This work represents a promising way to assess sleep-deprived fatigue, and our study provides a viable and efficient computational framework for user fatigue modeling in large-scale via social media.

\end{abstract}

\begin{IEEEkeywords}
Fatigue Prediction; Fatigue Modeling; Fatigue Analysis; Social Media; Selfies
\end{IEEEkeywords}

%
\IEEEpeerreviewmaketitle

\section{Introduction}

In modern societies, with growing pressures from life and work, fatigue becomes a common condition or even symptom for many people. Even if fatigue might be caused by illnesses, more of the population experience fatigue corresponding to situational factors or lifestyle such as sleep deprivation, depression, and anxiety \cite{jason2010fatigue}. According to \cite{jason2010fatigue}, long duration fatigue is usually caused by chronic illnesses, or psychiatric disorders. This kind of fatigue symptom is classified as pathological, and it may bring significant impairments to individual’s personal well-being. Another common type of fatigue is non-pathological, which is caused by short-term sleep deprivation, heavy labor, or flu-like illnesses. None-pathological fatigue is self-inflicted and has short duration, but its cause is comparatively less identifiable. 
Many researchers have been studying the etiology and classification of fatigue, as well as systematically assessing fatigue \cite{smets1995multidimensional}. However, those assessments are based on self-report or self-rating, which requires patient participation. This restricts the scale of study, and may also potentially carry biases. In our study, we concentrate on addressing sleep-deprived fatigue (which we call fatigue in the rest of the paper) by leveraging the state-of-the-art deep neural networks, and large-scale diverse data from social media. We believe analyzing social media posts and pictures offers the potential to provide a method for massive screening for individuals at risk for a range of poor health conditions. Instagram has been used to pick up signals when individuals use language that could be related to risky behaviors \cite{pang2015monitoring}. Tracking pictures can work in the same way to detect risks and allow for early intervention.
Therefore, we are particularly interested in:

\begin{itemize}
\item finding a new method to predict fatigue effectively and efficiently; and
\item applying this new method to analyze the fatigue of a massive number of users on the social media.
\end{itemize}

\begin{figure*}
\centering
\includegraphics[width=\linewidth]{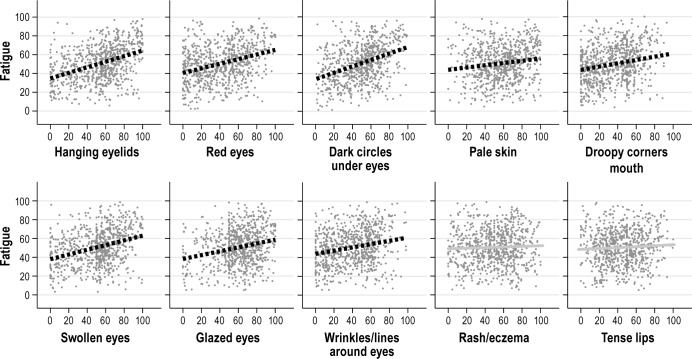}
\caption{Relationships between fatigue rates and the eight facial cues rates, reprinted from \cite{sundelin2013cues}. The first eight charts demonstrate that sleep-deprived fatigue correlates with the eight facial cues, whereas the last two charts show that sleep-deprived fatigue does not correlate with Rash/eczema and Tense lips.}
\label{fig:one}
\end{figure*}

With respect to our first goal, a research paper has quantitatively associated fatigue rate with human facial cues \cite{sundelin2013cues}. According to \cite{sundelin2013cues}, fatigue is heavily correlated with eight facial cues, including hanging eyelids, red eyes, dark circles under eyes, pale skin, droopy corner mouth, swollen eyes, glazed eyes, and wrinkles/lines around eyes. Figure \ref{fig:one}, which is reprinted from \cite{sundelin2013cues}, shows the correlations between the perceived fatigue rate and those facial cues. The rate from 0 to 100 indicates the degree of the facial cue from “not at all” to “very”.

Based on the correlation coefficients given in Figure \ref{fig:one}, we have constructed a combined estimator (explained in section 2.4) to measure the overall fatigue rate. We train a statistical model (explained in section 2) to predict the overall fatigue rate. With this model, given any selfie, we can predict the perceived fatigue rate within milliseconds.  

We apply our prediction method on all faces obtained from 10,480 Instagram users. For each face detected from use timelines, we first utilize the Face++ API to 1) infer its age, gender, race and facial landmarks; 2) identify user’s faces in his/her timeline via face recognition; and 3) use the facial landmarks to locate the interest areas that can indicate the facial cues on the face. These interest areas are sent to our models to produce the overall fatigue rate. 

We then analyze the fatigue rate at the user-level in terms of age, gender, race, and posting time.

Specifically, our main contributions include:

\begin{itemize}
\item We develop a new method to predict fatigue rate for a given face (e.g., selfie on social media). 
\item We apply our methodology at a large scale to study the population fatigue using selfies from social media, and discover the trends of sleep-deprived fatigue among different demographical groups.
\end{itemize}

\section{RELATED WORK}

Our work builds upon previous research on sleep condition, fatigue studies, and computer vision.\cite{sundelin2013cues} have found that the sleep-deprived fatigue rate is correlated with eight facial cues, and our study is based on the assumption that sleep-deprived fatigue is indeed mainly reflected by those eight facial cues, and the fatigue rate can somehow imply sleep conditions.
In terms of computer vision research, our work is related to face detection \cite{viola2001rapid}, gender, race and age identification \cite{han2014age}, facial landmarks location \cite{wang2017facial}, and face grouping \cite{zhao2003face}.

\section{MODEL CONSTRUCTION}
The main steps of the model construction are as follows:
\begin{enumerate}
\item Collect a training set of high-quality facial images
\item Have raters who possess prior clinical research experiences to rate the eight facial cues for each training face
\item Extract the feature descriptors from the training face
\item Train models with those feature descriptors and the facial cue rates.

\end{enumerate}

Briefly, for each given face, we need to check eight facial cues that are highly correlated with fatigue, as Figure 1 shows. To compute the fatigue rate in accordance with the linear regression estimators given by Figure 1, the rate for each facial cue is also necessary. Therefore, we have collaborated with the Clinical and Social Psychology Department at the University, which provides the ratings of the eight facial cues for each training face. Next, we extract the areas of interest from each training face, transform them into feature descriptor, and train a model to predict overall fatigue rate. 

\subsection{Training Data Collection}
An ideal training dataset would be a dataset that contains enough unique faces, some of which appear more fatigue than others, and some of which do not look fatigue at all. Moreover, for each unique face in the dataset, a few more facial images of that same face are also needed as references for the rating process (explained in section 2.2). Considering the expectations above, we chose the COLORFERET database as our training dataset. The COLORFERET database\footnote{https://www.nist.gov/itl/iad/image-group/color-feret-database} is sponsored by the Facial Recognition Technology (FERET) program of NIST\footnote{https://www.nist.gov}, with around 1000 unique faces in the entire database. For each face, more than five high quality facial imageries are provided. We have picked 964 faces out of the database as our training dataset. To show that our training dataset contains both fatigue and non-fatigue faces, we present several training face examples below in Figure \ref{fig:two}.

\begin{figure}
  \centering	
  \includegraphics[width=\linewidth]{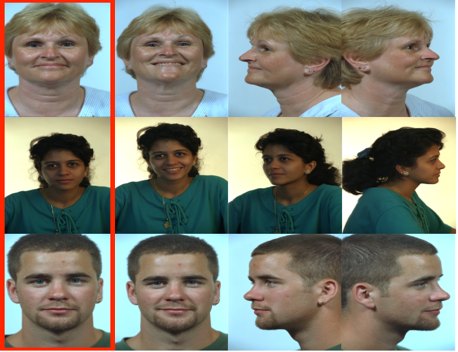}
  \caption{Examples of the training faces, note that the primary to-be-rated facial images are in the first column, while others are the reference images. Empirically, the training face in the second row appears most fatigue, while the faces in row one and row three appear less fatigue.}
  \label{fig:two}
\end{figure}

For facial detection and analysis tasks, we choose to employ the state-of-the-art face engine named Face++, which is an open-access face engine that provides services such as face detection, face recognition, demographical inference, and facial landmarks localization. The system is built with a CNN structure similar to the ImageNet structure as discussed in \cite{krizhevsky2012imagenet}, where five convolutional layers with max-pooling are connected with two fully connected layers and a softmax layer on top of them \cite{zhou2015naive}. Since the performance of Face++is reliable, we use it throughout our study. 

We use Face++ API to analyze the demographics of our training faces. Table \ref{table:one} shows the distribution of age, gender, and race in the training dataset. While one may suspect that temporary, fatigue-induced wrinkles could make the classifier label people as older than they are, we consider this a minor effect because 1) permanent wrinkles are more salient and thus more informative to the classifiers, and 2) all faces are processed in the same manner by Face++, which has been trained on a massive number of faces. 

\begin{table}[]
\centering
\caption{The age, gender, and race distribution in training dataset. The average confidences for gender and race predictions are 95.2\% and 90.5\% respectively.}
\label{table:one}
\begin{tabular}{cccccc}
\hline
\multicolumn{2}{c}{Age}                               & \multicolumn{2}{c}{Gender}                             & \multicolumn{2}{c}{Race}                    \\ \hline
\multicolumn{1}{c|}{0-20}  & \multicolumn{1}{c|}{140} & \multicolumn{1}{c|}{Male}   & \multicolumn{1}{c|}{658} & \multicolumn{1}{c|}{Asian}            & 148 \\
\multicolumn{1}{c|}{20-40} & \multicolumn{1}{c|}{607} & \multicolumn{1}{c|}{Female} & \multicolumn{1}{c|}{306} & \multicolumn{1}{c|}{African American} & 152 \\
\multicolumn{1}{c|}{40-60} & \multicolumn{1}{c|}{178} &                             & \multicolumn{1}{c|}{}    & \multicolumn{1}{c|}{Caucasian}        & 664 \\
\multicolumn{1}{c|}{60-80} & \multicolumn{1}{c|}{39}  &                             & \multicolumn{1}{c|}{}    & \multicolumn{1}{c|}{}                 &    \\ \hline
\end{tabular}
\end{table}

\subsection{Training Data Rating}
\label{section:2.2}
For each training face, there are eight facial cues, namely, hanging eyelids, red eyes, dark circles, pale skin, droopy corner mouth, swollen eyes, and wrinkles around eyes, to rate. With the purpose of making the rating as accurate and objective as possible, we have invited three experts with prior clinical research experiences to help us rate the training faces. All three raters have received an inter-judge agreement that describes the details about which area to scrutinize while rating a facial cue. 

Integers from 0 to 4 were used to indicate the rate of each facial cue, and 0 means \textit{not at all} whereas 4 means \textit{very}. In addition, we apply three techniques to keep the rating even more objective. 

First, we ask each of the three raters to rate eight facial cues for all 964 training faces, and the final rating for a facial cue of a face is calculated as the average of the ratings given by the three raters. 

Second, sometimes the rater may be influenced by the previous facial image while rating the current one. Such influence could be significant if the display order is the same for all three raters. Therefore, we randomize the display order of the training faces for each rater to minimize such influence. Third, while displaying a face, the actual to-be-rated facial image is displayed, along with four or more images of the same face as references. Given the average rate for each facial cue, we compute the overall fatigue rate for each training face through Equation \ref{eqn:01}. The following Figure \ref{fig:three} shows the overall fatigue rate distribution among the training faces.

\begin{figure}
  \centering	
  \includegraphics[width=\linewidth]{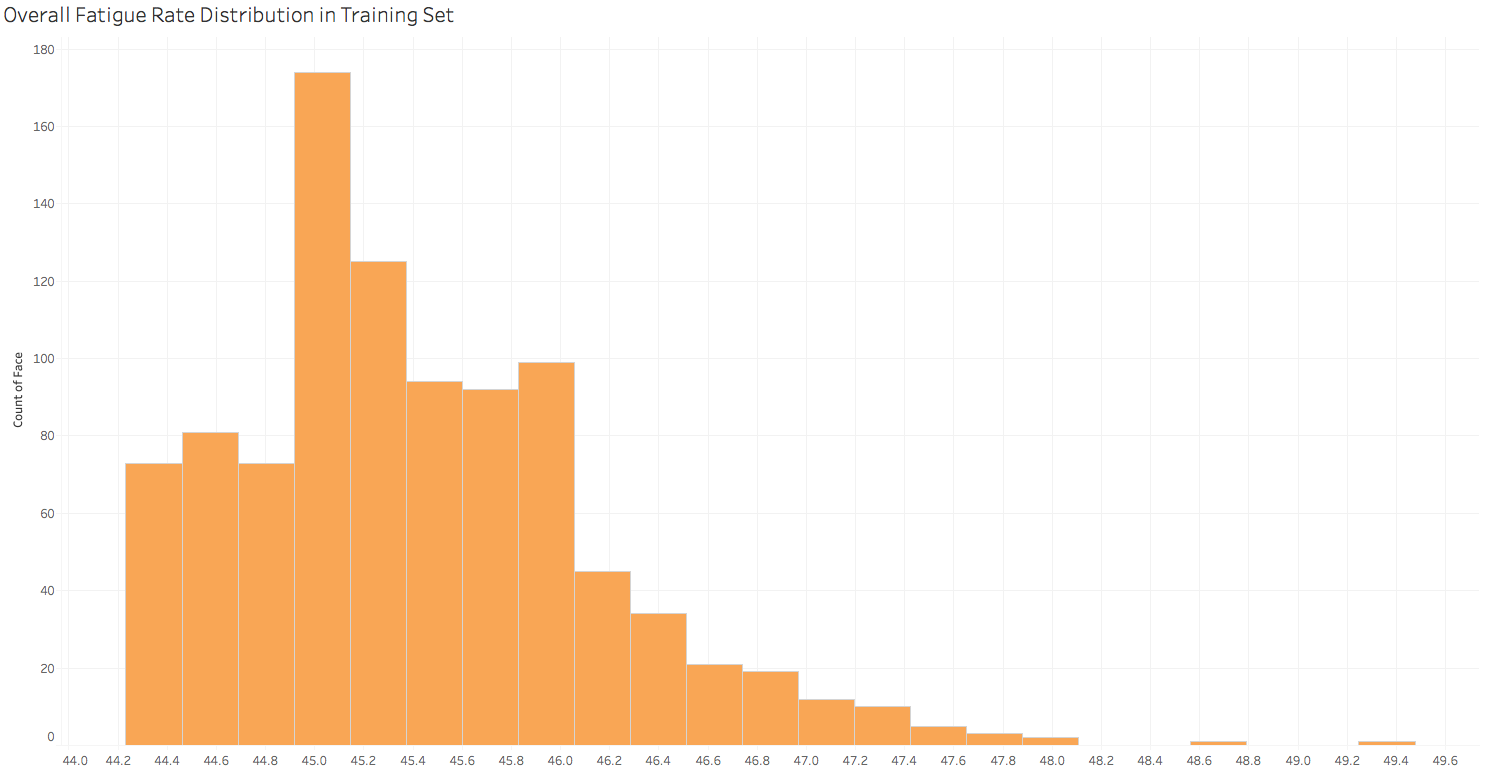}
  \caption{Normalized overall fatigue rate distribution among the training faces.}
  \label{fig:three}
\end{figure}

\subsection{Feature Extraction for Facial Images}

Six areas of interest are used to identify the eight facial cues of any given face. Left and right eye areas are examined to identify hanging eyelids, red eyes, swollen eyes, glazed eyes, and wrinkles/lines around eyes. The left and right eye bottom areas identify dark circle under eyes. The cheek area is for pale skin, and finally the mouth area is for droopy corner mouth. 

We apply a deep neural network, VGG16 \cite{simonyan2014very} with weights pre-trained on ImageNet, as our feature extractor. 

\begin{figure}
  \centering	
  \includegraphics[width=\linewidth]{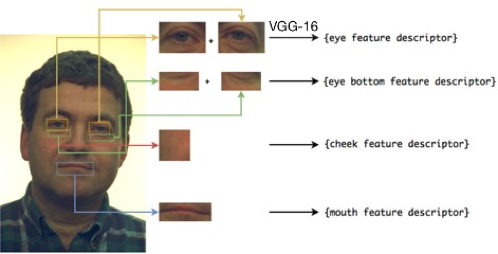}
  \caption{The feature extraction process, given a facial image, the feature extraction process will output four feature descriptors.}
  \label{fig:four}
\end{figure}

For each training face, we first call the Face++ API to obtain the facial landmarks of that face. We then crop the areas of interest from the original facial image accordingly. Subsequently, VGG16 will extract a feature descriptor for each resized cropped image. Note that for eye-related areas of interest, we concatenate the feature descriptors of the left and right eyes as a single feature descriptor, which we will call it eye feature descriptor; and we create the eye bottom feature descriptor in the same fashion. Figure \ref{fig:four} illustrates the entire feature extraction process.

\subsection{Model Training and Prediction}
\label{section:2.4}
For each training face, after obtaining four feature descriptors, we concatenate them into one feature vector.

Then, We fit our data by an ensemble of regression leaner. we train the ensemble learners with the feature vector and the ground truth overall fatigue rates. 

We perform Bayesian Optimization \cite{snoek2012practical} to optimize the hyper-parameters of our models. The optimization aims to locate the combination of 1) ensemble-aggregation method, 2) number of ensemble learning cycles, 3) learning rate for shrinkage, and 4) minimal number of leaf node observations that produces the least 5-fold cross validated RMSE. The following Table \ref{table:two} shows the best hyper-parameters for the ensemble learners found within 100 iterations, along with the RMSE rates. 

\begin{table}[]
\centering
\caption{Located hyper-parameters of the ensemble learners}
\label{table:two}
\begin{tabular}{{ccccc}}
\hline
Method  & Learning Cycles & Learn Rate & min Leaf Size & RMSE \\ \hline
LSBoost & 496             & 0.03       & 7             & 12.7
\\ \hline
\end{tabular}
\end{table}

Our final model produces a 0.28 10-fold cross validated SMAPE \footnote{$ sMAPE = \frac{100\%}{n}\sum_{t=1}^{n}\frac{|F_{t} - A_{t}|}{|F_{t}|+|A_{t}|} $} on the overall fatigue rates.

We locate the coordinates of each linear regression estimator (black dot line) in the charts given by Figure \ref{fig:one} to derive its mathematical expression. Since the ranges for the fatigue rate and each facial cue rate are equivalent, we therefore can derive a combined linear regression estimator to compute the overall fatigue rate by averaging those eight models as

\begin{equation}
\begin{split}
 y=0.037x_{1}+0.03x_{2}+0.041x_{3}+0.014x_{4}+ \\
 0.022x_{5}+0.033x_{6}+0.027x_{7}+0.024x_{8}+44.41 
\end{split}
\label{eqn:01} 
\end{equation}

This equation computes the overall fatigue rate for a given face.  $x_{1}, x_{2}, x_{3}, x_{4},x_{5}, x_{6},x_{7}$ and $x_{8}$ represent the rate of hanging eyelid, red eye, dark circle, pale skin, droopy corner mouth, swollen eye, glazed eye, and wrinkles, respectively.

\section{SOCIAL MEDIA SELFIE ANALYSIS}
In this section, we present how we apply our method to the faces collected from social media. We first describe how the user selfies are obtained, followed by how the faces within those selfies are used for fatigue prediction. 

\subsection{Selfie Collection}
\label{section:3.1}
\subsubsection{Collection}
There are nine keywords that people frequently tag when they are posting selfies on social media. They are ``\textit{\#selfie}'', ``\textit{\#me}'', ``\textit{\#happy}'', ``\textit{\#fun}'', ``\textit{\#smile}'', "\textit{\#nomakeup}'', ``\textit{\#friend}'',``\textit{\#family}'', ``\textit{\#fashion}'', and ``\textit{\#summer}''. We use those tags to search the photo posts on Instagram. For each post, we are able to acquire a uid (user Id), and such uid can be utilized to backtrack the uid owner’s timeline posts. Since we need enough samples from each user, timeline with less than 20 posts will be discarded. We collected nearly 10,480 users and 1-million posts from their timelines. 

\subsubsection{Low Quality Selfie Filtering}
\label{section:3.1.2}
Since the accuracy of our prediction depends on facial cues, the resolution of the selfie images and the quality of the face matter. In order to analyze faces of high quality only, we filter our collection of selfie images by considering three criteria: 1) Face eye and glass status, 2) blurriness of the image, and 3) face quality for facial recognition. 
For the first one, eye feature is one of our most important feature. Therefore, we need to make sure that we only assess faces with eyes open, and wearing no glasses. 
For any given face, there are five possible eye statuses, they are 

\begin{align*}
\textit{no\_glasses\_eye\_open},\textit{no\_glasses\_eye\_close}, \\
\textit{normal\_glasses\_eye\_open},\textit{normal\_glasses\_eye\_close}, \\
\textit{dark\_glasses}. 
\end{align*}

Face++ face detector will assess the confidence score for each status, we treat the one with the highest score as the eye status. 

Face++’s facial extensive landmark localization technology \cite{zhou2013extensive} is able to extract 83 points from a face, and each eye is described by 10 points. Thus, eye status can be identified easily by comparing the location of these points. Besides, Face++ detects eyewear using the methodology described in \cite{kumar2009attribute}. Figure 5 below demonstrates two examples for landmark localization and eyewear detection. 

\begin{figure}
  \centering
  \begin{subfigure}{0.5\linewidth}
  \centering
  \includegraphics[width=0.9\linewidth]{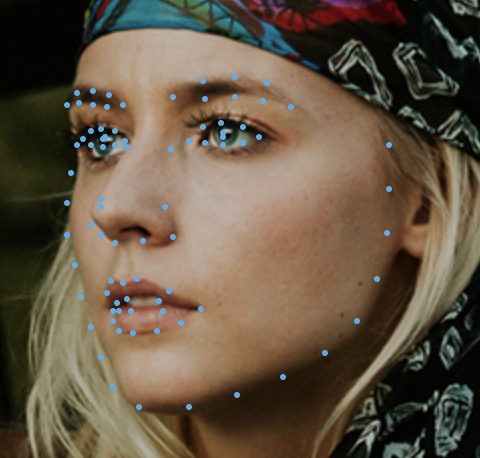}\hfill
  \caption{facial landmarks localization result}
  \label{fig:five:a}
  \end{subfigure}%
  \begin{subfigure}{0.5\linewidth}
  \centering
  \includegraphics[width=0.9\linewidth]{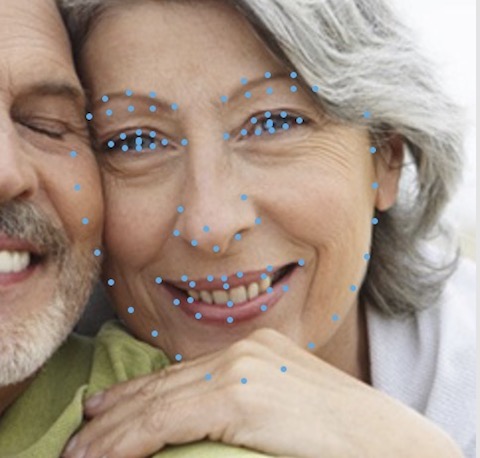}\hfill
  \caption{\textbf{Both Eye Status}: Open, no glass}
  \label{fig:five:b}
  \end{subfigure}
  \caption{Facial landmarks and attributes detection examples(images provided by Face++ demo)}
  \label{fig:five}
\end{figure}

We only keep the face whose eye status is \textit{no\_glasses\_eye\_open} for both left and right eyes. 
In terms of the remaining two factors, for any selfie, the detection result from Face++ includes two attributes: \textit{blur} and \textit{face\_quality}. The blur has two values, one is a floating-point number with 3 decimal places between 0 to 100, which represents the blurriness of the image; and another is the threshold that indicates whether the \textit{blur} has affected face recognition. In the same vein, \textit{face\_quality} indicates how suitable the image quality of face is, for face comparison. We only keep the face whose \textit{blur} value is lower than the threshold and \textit{face\_quality} value is higher than the threshold.

Those three criteria together help us select the faces which are most qualified for our study. After the fine-grained filtering, we obtain 9502 unique users, and 119,379 faces from their timeline posts.

\subsection{Selfie Processing and Prediction}
Even though those photo posts are tagged with the keywords mentioned in section \ref{section:3.1}, they still may contain zero number of face. Therefore, for any given photo post, our first step is face detection. Posts without faces were simply neglected.

Moreover, it is possible that photos in a user’s timeline contain not only his/her own faces, but also the faces of his/her friends, family members, and even strangers. Therefore, in a selfie, we consider the face with largest bounding box as the user’s face, based on the assumption that this face is closest to the camera; and thus, most likely to be the user. In addition, we also double confirm the face we are looking at belongs to the user by matching the gender and race throughout the user timeline. The following Figure \ref{fig:six} illustrates this process. 

\begin{figure}
  \centering	
  \includegraphics[width=\linewidth]{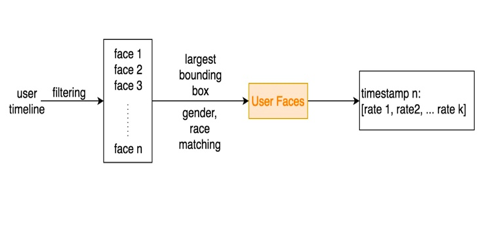}
  \caption{Procedure of timeline face detection, user identification, and overall fatigue rate prediction.}
  \label{fig:six}
\end{figure}

Among those high-quality faces, we identify 83039 user faces. We predict the overall fatigue rate for each of those user face, and then we group those faces by the timestamp component: weekday. More specifically, for each user, we put his/her faces posted on the same weekday into one set. The overall fatigue rate  for user in weekday  is computed as the average. Thus, for each weekday, we only include the same user once.

\section{EMPIRICAL TESTING}

The traditional way to conduct empirical testing on the effectiveness of our model could consume enormous time and labor, and it is not feasible to contact the users we collected from Instagram. Thus, we propose to validate our model by comparing the overall rates of faces from posts tagged with sleep-deprived related keywords, and good-sleep related keywords. We collect around 600 images for each group, and to reduce the sample biases, tags we included are: \textit{}{\#insomnia}, \textit{\#nosleep}, \textit{\#sleepdepreviation}, \textit{}{\#goodsleep}, \textit{\#freshmorning}, and \textit{\#tightsleep}. The images tagged with those keywords contain noise, so we manually remove the pictures without any face. We then apply the same procedure to filter faces, detect faces, and predict overall fatigue rates. The following Figure \ref{fig:seven} shows the bar chart of the mean overall fatigue rate from these two groups. 

\begin{figure}
  \centering	
  \includegraphics[width=\linewidth]{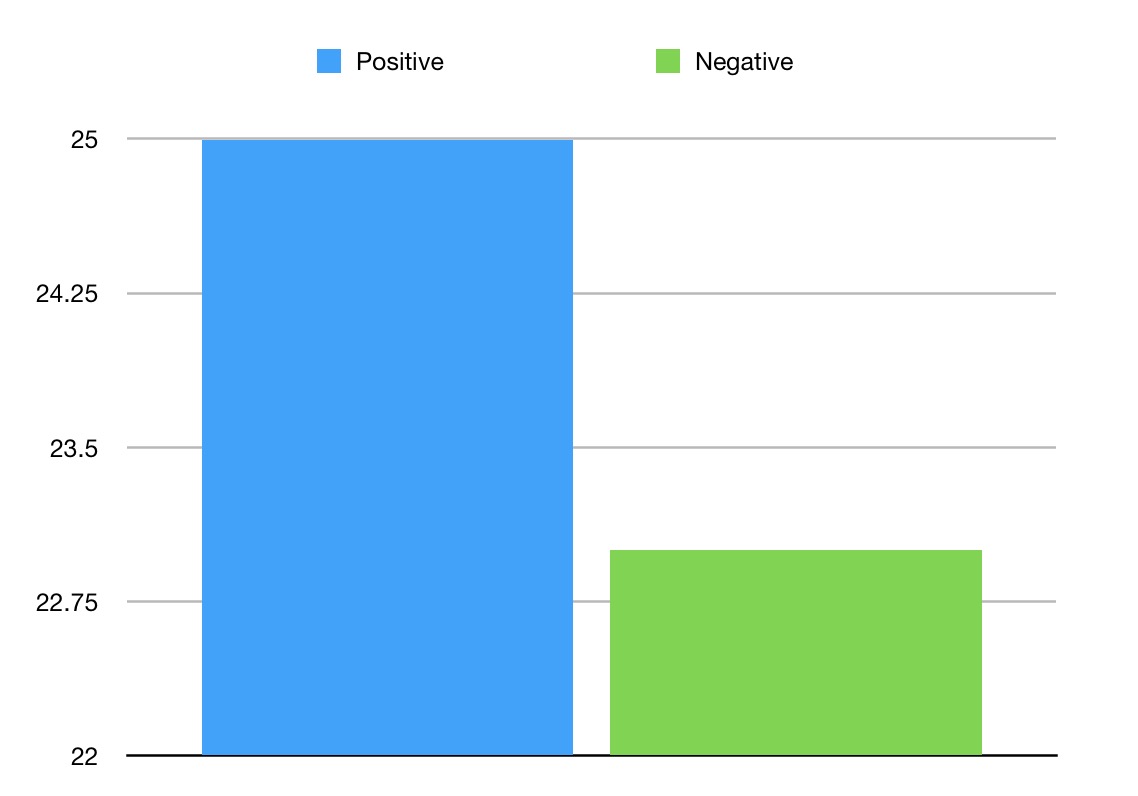}
  \caption{Mean overall fatigue rate value comparison between faces in two groups. The left bar represents the mean rate of faces tagged with keywords related to bad sleep.}
  \label{fig:seven}
\end{figure}

We further apply 2-sample t-test to compare the mean values between these two groups. The result rejects the null hypothesis that there is no difference between two-group means with a p-value of 0.0129; and the mean difference is between 1.050 to 8.433 with 95\% confidence level. 

\section{MAIN RESULTS}

The general demographic distributions of collected Instagram users are summarized in Table \ref{table:three}, Table \ref{table:four}, and Table \ref{table:five}. 

\begin{table}[]
\centering
\caption{Age Distribution}
\label{table:three}
\begin{tabular}{cccccccc}
\hline
0-10 & 10-20 & 20-30 & 30-40 & 40-50 & 50-60 & 60-70 & 70-80 \\ \hline
99   & 499   & 2172  & 4162  & 2124  & 372   & 64    & 10    \\ \hline
\end{tabular}
\end{table}

\begin{table}[]
\centering
\caption{Gender Distribution with average prediction confidence 94.6 \%}
\label{table:four}
\begin{tabular}{cc}
\hline
Male & Female \\ \hline
2994   & 6508   \\ \hline
\end{tabular}
\end{table}

\begin{table}[]
\centering
\caption{Race Distribution with average prediction confidence 86.2 \%}
\label{table:five}
\begin{tabular}{ccc}
\hline
Asian & African American & Caucasian \\ \hline
1897   & 539 &  7066 \\ \hline
\end{tabular}
\end{table}

\begin{figure}
  \centering	
  \includegraphics[width=\linewidth]{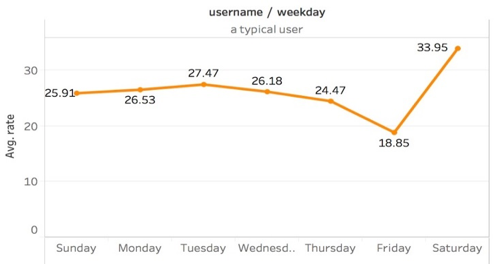}
  \caption{The average overall fatigue rates of a 41-year-old Female Caucasian throughout 7 days of a week.}
  \label{fig:eight}
\end{figure}

We analyze the overall fatigue rate on different weekdays among different age, gender, and ethnic groups. Figure \ref{fig:eight} shows a typical user’s average overall fatigue rate on seven weekdays. 

In the following subsections, we study the overall fatigue rates for different age, gender, and race groups in different weekdays in greater detail. Note that in our work, we always take age into consideration, because we think overall fatigue rates are more comparable within the same age group as users in the same age group tend to have relatively more similar life styles. We first show the overall fatigue rate trends on 7 weekdays among age groups; then, we add the gender and race into our analysis, respectively. We also report the statistical significance tests on multiple means comparison on each subsection.

\subsection{Age}
\begin{figure}
  \centering	
  \includegraphics[width=\linewidth]{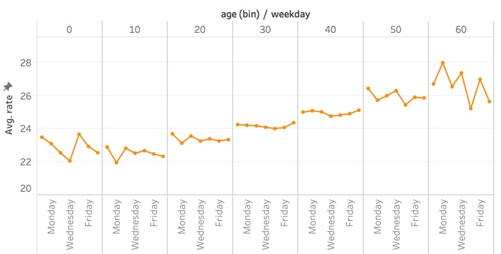}
  \caption{Overall fatigue rate trends on different weekdays over different age groups. (The weekdays from left to right are Sun., Mon., Tue., Wed., Thur., Fri., and Sat.)}
  \label{fig:nine}
\end{figure}

In this subsection, we examine the overall fatigue rate over 7 age groups on different weekdays. Since we only have 10 samples in the 70-80 age groups, we exclude this group from analysis. We report our results in Figure \ref{fig:nine}.

The results of statistical significance test suggest that people in the age group 10-20 tends to possess a higher overall fatigue rate on Tuesdays than on Fridays, and the difference in the mean overall fatigue rate is between 1.54 to 3.02 with 95\% confidence (p-value =0.0329). For the age group 30-40, people have a higher rate on Sundays than that on Tuesdays and Wednesdays, and the differences are between 0.51 to 0.99, and 0.62 to 1.12, respectively (p-values=0.04, 0.0042). Also, for this age group, the rate on Saturdays is higher than that on Wednesdays by a value between 0.61 to 1.11 (p-value=0.0075).

\subsection{Gender}

\begin{figure}
  \centering	
  \includegraphics[width=\linewidth]{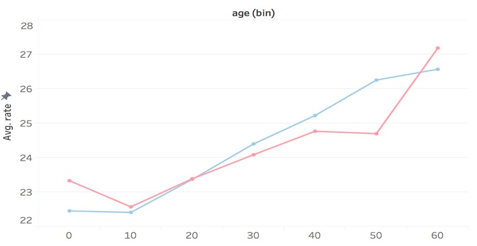}
  \caption{Overall fatigue rate trends on different weekdays over different age groups and genders.}
  \label{fig:ten}
\end{figure}

Figure \ref{fig:ten} shows the overall fatigue rates over two genders among age groups.

We found that in age group 30-40, the overall fatigue rate of males is higher than that of females, and the difference is between 0.43 to 0.61 with 95\% confidence (p-value=0.00). The 40-50 age group shares the same trend, the mean rate of males is 0.62 to 0.86 higher than that of the females (p-value=0.00).
Figure 12 takes the weekdays into consideration, and we found that on Fridays, males possess a higher rate than that of females, and we are 95\% confidence the difference is between 2.49 to 4.56. (p-value=0.0044).

\begin{figure}
  \centering	
  \includegraphics[width=\linewidth]{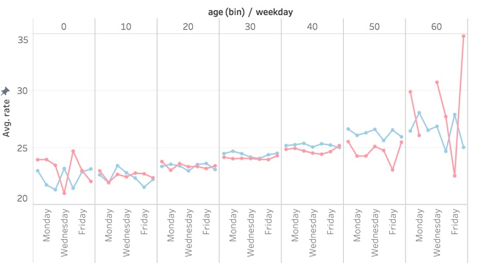}
  \caption{Overall fatigue rate trends on different weekdays over different age groups and genders. The pink and blue represent females and males, respectively}
  \label{fig:eleven}
\end{figure}

\subsection{Race}

\begin{figure}
  \centering	
  \includegraphics[width=\linewidth]{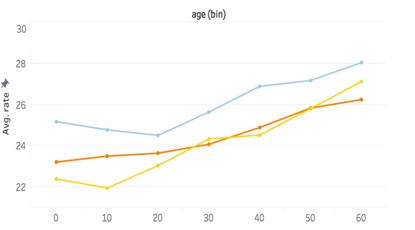}
  \caption{Overall fatigue rate trends over different age groups and races. The yellow, blue and orange represent Asian, African American, and Caucasian, respectively}
  \label{fig:twleve}
\end{figure}

In this subsection, we investigate the overall fatigue rate trends on different weekdays over race groups. Figure \ref{fig:twleve} shows the overall fatigue rates of different races among age groups.

We report the results of statistical significance test on race over different age groups in Table \ref{table:six}.

\begin{table}[]
\centering
\caption{Overall fatigue rate comparisons between races over different age groups. 1 stands for Asians, 2 stands for Caucasians, and 3 stands for African Americans. Note we only report the result whose p-value is less than 0.05.}
\label{table:six}
\begin{tabular}{ccc}
\hline
Race  & 95\% CI             & p-value \\ \hline
\multicolumn{3}{c}{Age Group 10 - 20} \\ \hline
1-2   & (-1.0121, -0.0246)  & 0.0369  \\
1-3   & (-2.0314, -0.4320)  & 0.0009  \\ \hline
\multicolumn{3}{c}{Age Group 20 - 30} \\ \hline
1-2   & (-0.7187, -0.4563)  & 0.0001  \\
1-3   & (-0.9605, -0.5691)  & 0.0019  \\ \hline
\multicolumn{3}{c}{Age Group 30 - 40} \\ \hline
1-2   & (-1.5654, -1.1341)  & 0.0000  \\
1-3   & (-1.2843, -0.6778)  & 0.0000  \\
2-3   & (0.1110, 0.6264)    & 0.0023  \\ \hline
\multicolumn{3}{c}{Age Group 40 - 50} \\ \hline
1-2   & (-1.1013, -0.4518)  & 0.0000  \\
1-3   & (0.5813, 1.4247)    & 0.0000  \\
2-3   & (1.4549, 2.1041)    & 0.0000  \\ \hline
\end{tabular}
\end{table}

\begin{figure}
  \centering	
  \includegraphics[width=\linewidth]{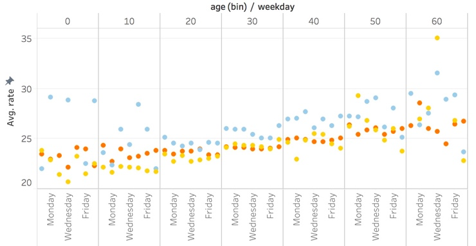}
  \caption{Overall fatigue rate trends on different weekdays over different age groups and races.}
  \label{fig:thirteen}
\end{figure}

Figure \ref{fig:thirteen} takes the weekday into consideration.

However, within the same age group, no statistical significant differences of mean overall fatigue rate are discovered among the racial groups.

\section{LIMITATION AND FUTURE WORK}

Our work is built on the assumptions that the sleep-deprived fatigue can indicate one’s sleep condition, and that sleep-deprived fatigue is associated the eight facial cues. However, it is possible that the fatigue appearance is caused by some other diseases such as systemic lupus erythematosus and rheumatoid arthritis \cite{swain2000fatigue} or a day of extremely heavy labor. Note that the use of social media selfies as the sensory data partially mitigates the above factors because people in those conditions are very unlikely to post selfies. In addition, some people may naturally appear more fatigue than others thus using fatigue to imply their sleep conditions could be biased. Therefore, in the future, we plan to develop more robust techniques to establish the face appearance baseline reference for a given individual before inferring his/her sleep condition. In addition, the overall fatigue rates of individuals who wear make-up can be underestimated. That is why we included \textit{\#nomakeup} as one of our keywords while retrieving selfies on social media. 
In addition, the resolution of the image, and the clearness of the face may also affect the prediction; however, we minimize this effect by refining our collection of social media selfies using three criteria described in section \ref{section:3.1.2} In addition, our empirical testing assumes that faces from Instagram posts tagged with sleep-deprived keywords indeed represent sleep-deprived individuals; but, this may not always be true. Our future work includes conducting empirical testing with human subjects beyond the social media.   
Lastly, due to the restriction in accessing the original data in \cite{sundelin2013cues}, Equation \ref{eqn:01} is solely derived by averaging those eight linear estimators presented in Figure \ref{fig:one}, which may lead to small inaccurate measurement of the impact each facial cue constitutes for the overall fatigue rate. Nonetheless, we believe the trends and distributions in our study will remain consistent and valid, especially given a significantly large number of face images is analyzed. 
During our prediction phase (explained in Section \ref{section:2.4}), the overall fatigue rate given by the combined estimator tends to be relatively low. This is because we have found out that in the training data rating step (explained in Section \ref{section:2.2}), the rates of facial cues from the three raters tend to be conservatively low, causing the overall fatigue rate to be low in the end. In the future work, we can correct such conservative ratings by giving more references and thus more confidence to the raters. 
Our current research primarily focuses on studying the fatigue rate with respect to demographical information. In the future, we may consider ways to identify the occupations of the face owners to investigate the distribution over different occupations. A recent study \cite{he2016using} has proposed a way to identify if a Twitter user is a college student or not. This method may help us unveil more interesting fatigue patterns of the student population. Moreover, we are also interested in identifying the geo-location of the users such that we can examine the fatigue distribution over different geographical areas. With these extensions, we may be able to verify that indeed people who live in regions where work and life environments are more intense and competitive, such as San Francisco and New York City, tend to appear more fatigue. 

\section{CONCLUSION}

We have developed a new method to gauge a face’s overall fatigue rate. This leads to a data-driven methodology to include a massive number of users on social media to obtain the fatigue distributions with respect to age, gender, and race in conjunction of time component. Moreover, the proposed approach is an example of using social networks for user behavior modeling, which could be further extended to risky behavior modeling. Some of our findings are beyond those reported in the literature, e.g., those related to the interplays of age, gender and race, pointing to the potential to discover factors that the empirical studies have overlooked.  Our work is in the same vein as \cite{pang2015monitoring}\cite{abdullah2015collective}\cite{wu2016effect}, and such social media-driven methods are expected to find more successes in computational sociology and psychology. 

\section*{Acknowledgment}
We thank the support of New York State through the Goergen Institute for Data Science, and our corporate research sponsors Xerox and VisualDX.

\bibliographystyle{IEEEtranBST/IEEEtran}
\bibliography{IEEEtranBST/IEEEexample}
\end{document}